\documentclass{article}

\PassOptionsToPackage{numbers, compress}{natbib}

\usepackage{subcaption}
\usepackage{graphicx}
\usepackage{amsmath}
\usepackage{multirow}
\usepackage{longtable}

\usepackage[dblblindworkshop, final]{neurips_2026}

\workshoptitle{TS-LIMITS}

\usepackage[utf8]{inputenc} 
\usepackage[T1]{fontenc}    
\usepackage{hyperref}       
\usepackage{cleveref}
\usepackage{url}            
\usepackage{booktabs}       
\usepackage{amsfonts}       
\usepackage{nicefrac}       
\usepackage{microtype}      
\usepackage{xcolor}

\title{Detecting Brain Cancer Progression Through Signatures of Wearable Accelerometry Data}

\author{
  Rishit Das \\
  Department of Computing\\
  Imperial College London\\
  \And
  Barbora Barancikova\\
  Department of Computing\\
  Imperial College London \\
  \And
  Matt Williams \\
  Department of Surgery \& Cancer\\
  Imperial College London \\
}

\begin{document}

\maketitle

\begin{abstract}
  Glioblastoma and other high-grade gliomas (HGG) have a poor prognosis and are associated with progressive functional decline. The standard of care to assess progressive disease is contrast-enhanced MRI, which is expensive, claustrophobic, and performed only once every 2-3 months. Wearable accelerometers offer a non-invasive alternative for continuously monitoring patient function between clinic visits. However, accelerometry data streams are typically noisy and highly patient-specific, making conventional time series methods and summary statistics poorly suited to detecting clinically meaningful change. This work aims to classify stable versus progressive disease states by applying rough path signatures for feature extraction from wrist-worn accelerometer data. We convert accelerometer recordings from HGG patients enrolled in the BrainWear study into walking-path segments and use signature kernel maximum mean discrepancy (SigMMD) to quantify change in patients' walking behaviour. We overcome the challenge of inter-patient heterogeneity by adjusting for the natural variation in a patient's walking during stable tumour periods. This patient-specific normalisation substantially improves separation, yielding a candidate biomarker of tumour progression. Our results provide proof-of-concept evidence that signature-based analysis of wearable accelerometry can distinguish stable from progressive disease states, motivating larger longitudinal studies of continuous wearable monitoring as a digital biomarker in neuro-oncology.

\end{abstract}

\section{Introduction}

Brain tumours are the leading cause of cancer death in individuals under 40~\cite{gbmdumba}. The most common brain tumour in adults, Glioblastoma (GBM), has a median survival of 15 months with aggressive treatment (surgery, radiotherapy, and chemotherapy)~\cite{Stupp2005}. Patients with a brain tumour present a range of functional deficits (e.g. weakness, visual problems), either at diagnosis, as a result of treatment, or due to progressive disease. Patients often develop additional symptoms as their disease progresses, and these may precede clinical detection of progressive disease.\par               

Patients typically develop progressive disease at 4-6 months after diagnosis. Disease progression is determined using contrast-enhanced MRI. However, MRI scans are expensive, time-consuming, claustrophobic, and often unfeasible in patients with pacemakers or shrapnel; in addition, an MRI scan can only demonstrate disease once it has progressed. Recent developments in wearable technologies have made it possible to continuously track patient activity, and there is an open question about whether we can use wearables to detect changes in patient function caused by recurrence or progression before they are detected by MRI scans.\par Although attractive, wearables pose significant technical challenges. Specifically, patients have different patterns and levels of activity, which change in different ways as their disease progresses based on diverse tumour locations and patterns of spread. In addition, analysis of such data remains challenging. For instance, it is unclear what the normal levels are, how these change with treatment, or what is an optimal criterion for detecting a significant change in activity. Also, because progression labels are often very limited, it is useful to extract training-free features from accelerometry data to reduce the dimensionality of the problem before applying machine learning. A simple and commonly used feature of this kind is step count, with prior work examining the prognostic impact of timed walk distances in glioma~\cite{jones_assessment_2009}. However, step count is a restrictive measure, and we hypothesise that it omits useful information about the temporal structure of walking, motivating our use of path signatures. If accelerometer-derived features demonstrate predictive value, they may support earlier detection of functional decline, improve monitoring between formal assessments, and complement conventional clinical evaluations. \par

Path signatures~\citep{lyons2007differential} and signature kernels~\citep{salvi2021signature} are tools from rough path theory, a branch of mathematics developed to study complex paths and data streams. They allow meaningful information about the ``shape'' of multi-dimensional time series to be extracted in a training-free manner. Signatures have recently gained popularity in the machine learning literature because they act as natural features for downstream models involving sequential data, including classifiers for distinguishing mental disorders from streams of self-reported mood~\citep{perez2018signature}, human action detection~\citep{yang2022developing}, Chinese character recognition~\citep{xie2017learning}, and generative models for time series~\citep{barancikova2025sigdiffusions}. Signature kernels measure the similarity between individual time series by comparing their path signatures; signature maximum mean discrepancy (SigMMD) uses MMD~\cite{gretton2012kernel} with a signature kernel to quantify differences between two distributions of time series through their signatures, with larger values indicating greater discrepancy.

In this work, we present a proof-of-concept study applying personalised signature kernel MMD to wrist-worn accelerometer data from patients with high-grade glioma, in order to distinguish stable from progressive disease using walking behaviour alone. For each clinic window, we use SigMMD to quantify changes in walking relative to an initial patient-specific baseline. To account for natural inter-patient variation, we then normalise these changes by the level of walking-pattern variation observed during each patient’s stable disease, as described in \Cref{sec:methods}. We use this personalised SigMMD as a one-dimensional decision score for disease progression classification in \Cref{sec:experiments}, where it achieves better discrimination between stable and progressive disease than analogous step-count-based measures.

\section{Background}\label{sec:bg}
\subsection{Rough Path Signatures}\label{secA1}

Rough path theory provides tools for representing time-ordered data in a way that captures interactions between different channels over time. In this work, each walking segment is represented as a continuous path $x : [0,1] \to \mathbb{R}^d$, obtained by linearly interpolating the observed time series values and reparameterising the segment to the unit interval. Here, $d$ denotes the number of channels: $d=4$ for the original triaxial representation augmented with time and $d=2$ for the ENMO representation, in which the three accelerometer channels are transformed into a single Euclidean Norm Minus One (ENMO) channel.

The signature $S(x)$ of $x$ is defined as the following sequence of tensors of increasing order
\begin{equation}
S(x) = \left(1, S_1(x), S_2(x), \ldots\right).
\end{equation}
Here, the \(k\)\textsuperscript{th} term is defined as a \(k\)-fold tensor of iterated integrals,
\begin{equation}
S_k(x)=\int_{0\leq t_{1}<\cdots<t_{k}\leq 1} dx_{t_{1}} \otimes \cdots \otimes dx_{t_{k}} \quad \text{for } k \geq 1,
\end{equation}
where $\otimes$ denotes the tensor product. These tensors are often referred to as \textit{levels}. For example, the tensor at level one is a \(d\)-dimensional vector, level two is a \(d\times d\) matrix, and so on.

Equivalently, one can write $S_k(x)$ by flattening the tensor structure and indexing each tensor entry as
\begin{equation*}
S_k(x)
= \big( S_k(x)^{i_1,\ldots,i_k} : i_1,\ldots,i_k \in \{1,\ldots,d\} \big),
\end{equation*}
where $S_k(x)^{i_1,\ldots,i_k}$, for a set of indices $i_1,\ldots,i_k \in \{1,\ldots,d\}$, is the $k$-fold iterated integral
\begin{equation*}
S_k(x)^{i_1,\ldots,i_k}
=
\int_{0\leq t_1<\cdots<t_k\leq 1}
dx^{i_1}_{t_1}\cdots dx^{i_k}_{t_k}.
\end{equation*}

\subsection{Signature Kernel MMD}\label{MMD}

Let $P$ and $Q$ be two probability distributions defined over a domain $\mathcal{X}$. 
Take a feature map $\phi : \mathcal{X} \rightarrow \mathcal{H}$ such that it maps 
elements from the original space into a reproducing kernel Hilbert space (RKHS), denoted as $\mathcal{H}$. The mean embedding of the distribution $P$ in $\mathcal{H}$ is defined as the expected value of its mapped features
\begin{equation*}
    \mu_P = \mathbb{E}_{X \sim P}[\phi(X)].
\end{equation*}

Maximum Mean Discrepancy (MMD)~\cite{gretton2012kernel} measures the distance between $P$ and $Q$ by calculating the distance between their 
respective mean embeddings, $\mu_P$ and $\mu_Q$, in the RKHS. By definition, each RKHS is associated with a unique positive-definite kernel $k$, which can be expressed in terms of the feature map $\phi$ as the inner product
\[
k(x,y)=\langle \phi(x),\phi(y)\rangle_{\mathcal H}.
\]
MMD can then be calculated in terms of the kernel as
\begin{align*}
\mathrm{MMD}^2(P,Q)
&= \|\mu_P-\mu_Q\|_{\mathcal{H}}^2 \\
&= \langle \mu_P,\mu_P\rangle_{\mathcal{H}}
 + \langle \mu_Q,\mu_Q\rangle_{\mathcal{H}}
 - 2\langle \mu_P,\mu_Q\rangle_{\mathcal{H}} \\
&= \mathbb{E}_{x,x' \sim P}\left[k(x,x')\right]
 + \mathbb{E}_{y,y' \sim Q}\left[k(y,y')\right]
 - 2\mathbb{E}_{x \sim P,\, y \sim Q}\left[k(x,y)\right].
\end{align*}

For finite sets of samples from $P$ and $Q$, we replace the expectations above with their empirical estimates. In this study, we use the path signature as our feature map, $\phi(x)=S(x)$. The corresponding signature kernel, i.e., the inner product between the signatures of two paths, can be computed efficiently without explicitly calculating the signatures~\citep{salvi2021signature}. We use SigMMD to quantify how different two collections of walking paths are, based on their temporal structure and cross-channel interactions. 

\section{Methods}\label{sec:methods}
In this section, we describe the BrainWear dataset, and the design of our training-free personalised SigMMD feature, which we later use as input for disease progression classification.
\subsection{Patient Population and Data Preprocessing} \label{subsec:Preproc}

The BrainWear study~\cite{10.1371/journal.pone.0285399} was a single-centre phase 2 feasibility study exploring the practicality of collecting data on HGG patients from a wrist-worn accelerometer. The study also tracked every patient's tumour progression as \textit{stable} or \textit{progressive} during their clinic visits. We use these labels as the ground truth for our modelling. The complete study dataset, which extends the cohort reported in \cite{10.1371/journal.pone.0285399}, comprises of 78 enrolled patients, with labels from 114 clinic visits, of which 70 were \textit{progressive} and 44 were \textit{stable}. The median age of the patients was 58 years, and 62.8\% were male.

To categorise the raw accelerometer data, we applied the UK Biobank activity classification pipeline~\cite{willetts2018statistical}, retaining only segments labelled as \textit{walking}. Contiguous \textit{walking} samples were considered to be part of the same walking bout. Every walking bout was represented as a three-dimensional time series using the triaxial accelerometer channels. Each channel was band-pass filtered independently using a fourth-order Butterworth filter with lower and upper cut-off frequencies of 0.5\,Hz and 10.0\,Hz, respectively. Walking bouts shorter than 15\,seconds were discarded, and the remaining filtered bouts were downsampled from 25\,Hz to 5\,Hz for computational efficiency. The bouts were then divided into fixed-length 15\,second segments, giving paths with 75 time points each. The triaxial path representation was transformed into ENMO, yielding a one-dimensional time series of acceleration magnitude, augmented with an additional time channel increasing linearly from 0 to 1 over the 15\,second interval. Near-flat paths (defined as paths where the standard deviation across accelerometer channels was less than $10^{-6}$) were discarded as likely measurement artefacts. All valid walking paths from each calendar day were collected into a set. 

\subsection{Personalised SigMMD} \label{sec:personalisedSigMMD}
With the preprocessed 15 second samples of each patient's walking over time, we aim to capture long-term change in patients' walking patterns, as we hypothesise they might be indicative of disease progression. For this, we use the first few days of recorded walking as a \textit{baseline set} to establish the patients' typical walk, which we then compare to the walking samples from each of the subsequent days in the \textit{comparison set} using SigMMD. The full methodology is described below.

\paragraph{Baseline Set} \label{subsec:Baseline}
The time period between the previous clinic visit (or the first day of available data, if there is no previous clinic visit) and the current clinic visit is defined as the \textit{clinic window}. We constructed per-patient \textit{baselines} to represent each patient's usual walking behaviour at the start of each \textit{clinic window}. Because accelerometer wear varied between patients and valid walking data was not available every day, the baseline was constructed dynamically according to data availability. Let the ordered days within a clinic window be indexed by $\{1,\ldots,N\}$, where day $N$ is the clinic visit date. The baseline period started on day 1. We first considered an initial baseline period of 14 calendar days, or up to the day before the clinic visit if the clinic window was shorter. All days within this period containing valid walking bouts were included in the baseline. If fewer than 7 days contained valid walking bouts, the baseline period was extended forward one calendar day at a time until either 7 days containing valid walking bouts were found or the day before the clinic visit was reached. Clinic visits for which fewer than 7 valid baseline days were available were excluded from analysis. We normalised the baseline and subsequent comparison sets using the mean path norm from the baseline set. The  baseline construction was repeated for each patient and clinic window.

\paragraph{Comparison Sets}

After constructing the baseline, we assigned each subsequent day in the clinic window to the \textit{comparison period}, the period beginning on the day after the final baseline day and ending on the clinic visit date. Every day with valid walking paths in this interval formed a comparison set. Window coverage was defined as the proportion of calendar days in the comparison period that contained valid data, and clinic visits with window coverage below 50\% were excluded. For each clinic window, daily path data from each day was grouped into comparison sets. Let b be the last day of the baseline set. Then $\{b + 1, b + 2, \ldots, N\}$ is the set of all days in the comparison period. We define the \textit{comparison sets} as a collection of sets $\mathcal{S} = \{S_{b+1}, S_{b+2}, \ldots, S_{N}\}$ where each set $S_k$ contains walking paths from day $k$. 

\paragraph{Quantifying change in walking}

\begin{figure}[t]
\centering
\includegraphics[width=0.6\textwidth]{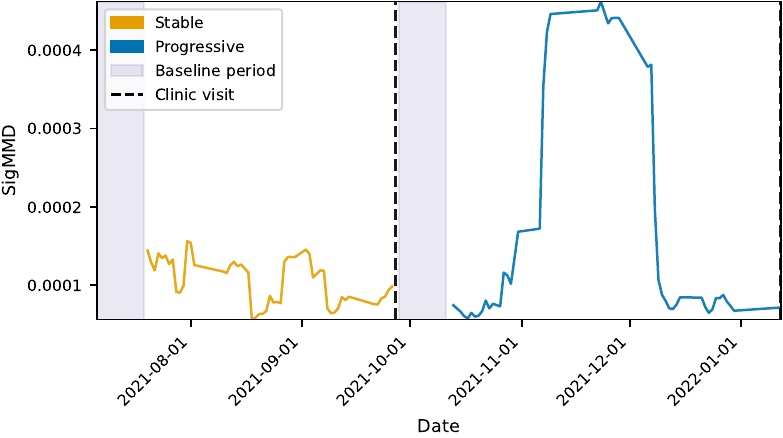}
\caption{Daily SigMMD values for a selected patient over two clinic windows. The shaded region shows the baseline period and the dashed line is the clinic visit date. The progressive window shows higher SigMMD values compared to the stable window.}\label{fig:sig_mmd_line}
\end{figure}

We treated each comparison set $S_k$ as an independent sample set and compared it with the baseline set $B$ for every clinic window. We quantified the change in patient's walking over time as $\mathrm{SigMMD}(B, S_k).$ This produced a time series of SigMMD values for each clinic window, with $N - b$ data points. SigMMD values for one patient across stable and progressive clinic windows are shown in \Cref{fig:sig_mmd_line}. To obtain a scalar summary for subsequent analysis, we computed the mean SigMMD across all comparison sets. Thus, each clinic window produced one raw mean SigMMD value, which was used as a feature for tumour progression classification. 

Although larger SigMMD values indicate greater deviation from the patient's baseline walking pattern, some deviation is expected even during stable disease due to patient-specific mobility patterns. Since the aim was to capture unusually large relative increases in SigMMD during disease progression, each clinic visit was normalised by the patient's mean SigMMD during stable visits. We refer to this ratio as \textit{personalised} SigMMD. Let $\mathcal{J}^{\mathrm{stable}}$ denote the set of a patient's clinic visits labelled as \textit{stable}. The stable reference value was defined as 
\begin{equation*} 
\mu^{\mathrm{stable}} = \frac{1}{|\mathcal{J}^{\mathrm{stable}}|} \sum_{j \in \mathcal{J}^{\mathrm{stable}}} \overline{m}_j, \end{equation*} 

where $\overline{m}_j$ denotes the mean SigMMD for clinic visit $j$. 

For each clinic visit $j$ from the same patient, the \textit{personalised} SigMMD was then defined as
\begin{equation*} 
r_j = \frac{\overline{m}_j}{\mu^{\mathrm{stable}}}. 
\end{equation*}

By construction, the ratios of a patient's stable visits average to 1, and a patient with a single stable visit has $r = 1$ for that visit. The evidence therefore comes mainly from the progressive visits, where $r > 1$ means that the walking pattern deviated more from its baseline than it did during the same patient's stable disease.

We use the personalised SigMMD ratio as a threshold biomarker, producing a one-parameter linear classifier model, i.e., a continuous decision score. Leave-one-patient-out (LOPO) cross-validation was used to evaluate model performance. For each training fold, a receiver operating characteristic (ROC) analysis was performed using the training patients only. The classification threshold was then selected by maximising the Youden's index, $J$. The threshold corresponding to the maximal value $J$ was taken as the fold-specific optimal operating point. After estimating the optimal threshold from the training set, this threshold was applied unchanged to all visits from the held-out patient. Each held-out visit was classified as \textit{progressive} if its personalised SigMMD ratio was greater than or equal to the threshold, and \textit{stable} otherwise. After completing all LOPO iterations, predictions from the held-out patients were aggregated across folds, with 95\% confidence intervals estimated from 1000 patient-level bootstrap resamples of the predictions.

\section{Experiments}\label{sec:experiments}
After preprocessing the data as described in \Cref{sec:methods}, we identified 21 patients with 36 clinic visits with sufficient activity data to generate SigMMDs. Further, 10 patients only had progressive clinic visits, leaving us unable to establish a stable baseline to normalise against for personalised SigMMD generation, resulting in a final cohort of 11 patients and 26 clinic visits. Of these, 11 clinic visits were labelled as progressive disease while 15 were labelled as stable. We analysed 3,030 hours of walking accelerometry data from these patients to obtain the following results.

\paragraph{Accelerometry signatures vary across activity}\label{subsec:activitydiff}

\begin{figure}[!ht]
\centering
\begin{subfigure}[t]{0.24\textwidth}
    \centering
    \includegraphics[width=\linewidth]{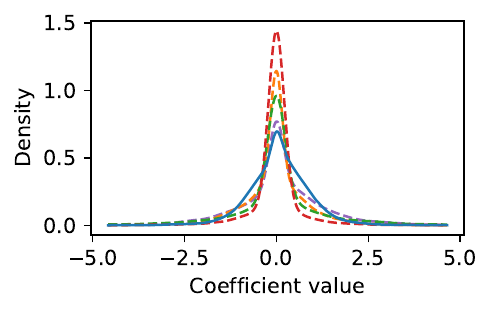}
    \caption{}
\end{subfigure}
\begin{subfigure}[t]{0.24\textwidth}
    \centering
    \includegraphics[width=\linewidth]{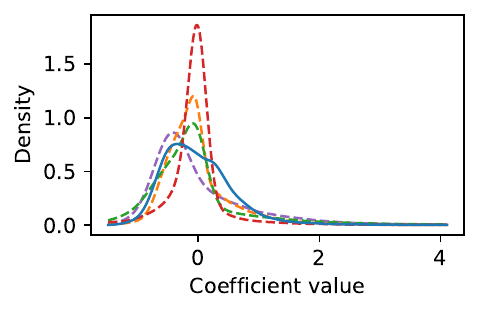}
    \caption{}
\end{subfigure}
\begin{subfigure}[t]{0.24\textwidth}
    \centering
    \includegraphics[width=\linewidth]{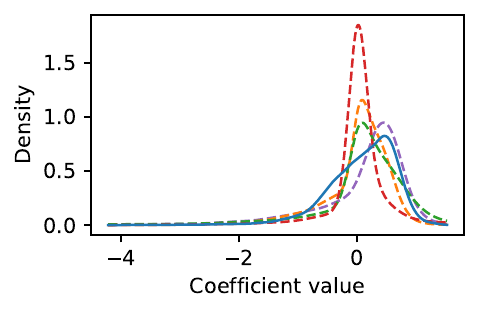}
    \caption{}
\end{subfigure}
\begin{subfigure}[t]{0.24\textwidth}
    \centering
    \includegraphics[width=\linewidth]{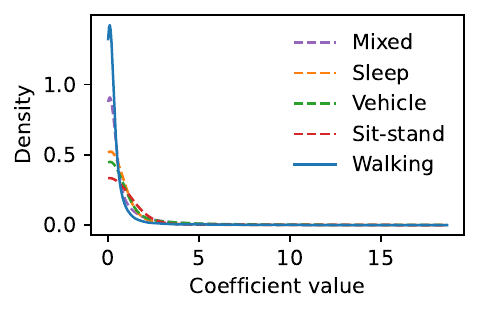}
    \caption{}
\end{subfigure}
\caption{Activity label distributions across signature coefficients: (a) $S_1(x)^2$, (b) $S_2(x)^{1,2}$, (c) $S_2(x)^{2,1}$, (d) $S_2(x)^{2,2}$, with 1,000 samples per label. Activities have distinct signature coefficient distributions. Specifically, Walking (in blue) shows a different distribution than other activities.}
\label{fig:signature_distribution}
\end{figure}

First, to confirm that path signatures provide a meaningful representation of movement from accelerometry data, we examined whether signature coefficients differed across activity types identified by the UK Biobank pipeline. We considered signature coefficients up to level 2, with representative distributions shown in \Cref{fig:signature_distribution}. For each activity label, we sampled 1,000 15 second accelerometer segments uniformly at random from across the patient cohort and applied two-sample Kolmogorov--Smirnov (KS) tests~\cite{kolmogorov1933sulla,smirnov1939estimation} to compare the distribution of each non-trivial signature coefficient between walking and the other activity classes. The signature coefficients consistently showed distinct distributions across activity types, supporting their use for characterising movement from accelerometry data. See \Cref{subsec:ksResults} for full results.

\paragraph{Personalised SigMMD is higher at progressive visits}\label{subsec:persResults}

\begin{figure}[ht]
\centering
\begin{subfigure}{0.4\textwidth}
\centering
\includegraphics[width=0.9\linewidth]{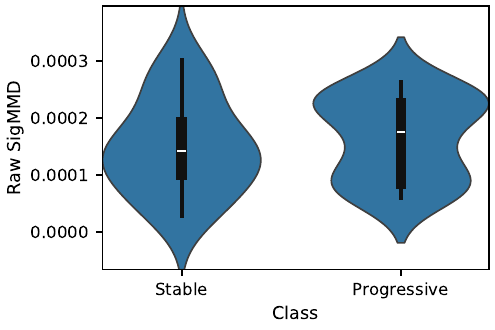}
\caption{Raw mean SigMMD.}
\end{subfigure}
\begin{subfigure}{0.4\textwidth}
\centering
\includegraphics[width=0.9\linewidth]{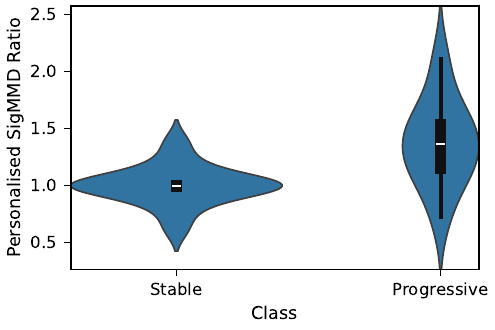}
\caption{Personalised SigMMD ratio.}
\end{subfigure}

\caption{Distributions of raw mean SigMMD values and personalised SigMMD ratios for stable and progressive-disease visits. Raw mean SigMMD values show substantial overlap between the two groups, whereas patient-specific normalisation reveals a tendency toward higher personalised SigMMD ratios among the progressive visits.}
\label{fig:violin_plots}
\end{figure}

We examined the effect of patient-specific normalisation by comparing the raw mean SigMMD values with the corresponding personalised SigMMD ratios described in \Cref{sec:personalisedSigMMD}. The distributions of both features are shown in \Cref{fig:violin_plots}. Raw mean SigMMD values showed substantial overlap between stable and progressive-disease clinic windows, indicating that absolute SigMMD values alone provide limited separation between the two classes. We tested whether personalised ratios are higher at progressive visits than at stable visits with a within-patient permutation test. We used the Mann-Whitney U statistic and obtained its null distribution by exhaustively permuting the stable and progressive labels within each patient, recomputing the stable reference and the ratios for every permutation. The reported p-value is the fraction of permutations with a statistic at least as large as the observed one. Personalised ratios were higher at progressive than at stable visits (p-value = 0.016 over 4,320 permutations).

\begin{figure}[ht]
\centering
\begin{subfigure}{0.48\textwidth}
    \centering
    \includegraphics[width=0.7\linewidth]{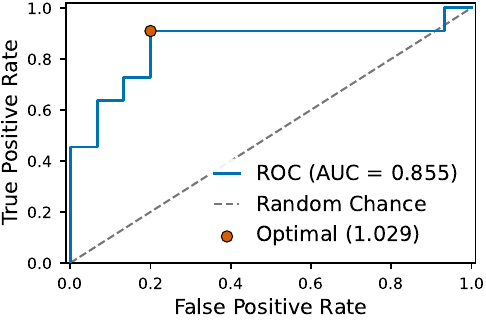}
    \caption{ROC curve generated from the personalised SigMMD ratio thresholds.}
    \label{fig:sub1}
\end{subfigure}
\hfill
\begin{subfigure}{0.48\textwidth}
    \centering
    \includegraphics[width=0.7\linewidth]{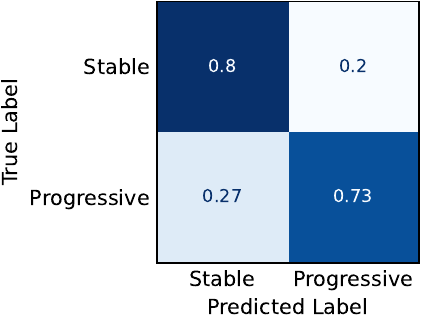}
    \caption{Confusion matrix generated by Leave One Patient Out Classification.}
    \label{fig:sub2}
\end{subfigure}
\caption{Performance evaluation of the personalised SigMMD classifier.}
\label{fig:conf_mat}
\end{figure}

We next evaluated the personalised SigMMD ratio as a marker of progressive disease using the LOPO procedure and bootstrap confidence intervals described in \Cref{sec:personalisedSigMMD}. The resulting ROC curve had an area under the curve (AUC) of $0.8545$ (\Cref{fig:sub1}) and the corresponding LOPO classification results are summarised by the confusion matrix in \Cref{fig:sub2}. Full performance metrics are reported in \Cref{tab:classification_metrics}. Repeating the analysis with the triaxial representation gave an AUC of 0.7455 (\Cref{sec:triaxial}).

\begin{table}[!ht]
\caption{LOPO classification performance with 95\% confidence intervals.}
\centering
\resizebox{\textwidth}{!}{
\begin{tabular}{lcccc}
\toprule
\textbf{Model} & \textbf{AUC}&\textbf{Sensitivity} & \textbf{Specificity} & \textbf{Precision}\\

\midrule
Ours (Personalised Step Count - Ablation)&0.4424 (0.1871-0.7620) &0.3636 (0.1111-0.6250) &0.8000 (0.6471-0.9375) &0.5714 (0.2000-0.8889) \\
60 Day Step Count Benchmark&0.4242 (0.1905-0.6727) &0.0909 (0.0000-0.2857) &0.8667 (0.6922-1.0000) &0.3333 (0.0000-1.0000) \\
\textbf{Ours (Personalised SigMMD)} &\textbf{0.8545 (0.6600-0.9925)}& \textbf{0.7272 (0.3636-1.0000)}  & \textbf{0.8000 (0.6471-1.0000)} & \textbf{0.7272 (0.4000-1.0000)}\\
\bottomrule
\end{tabular}
}
\label{tab:classification_metrics}
\end{table}

\paragraph{Step counts do not separate stable from progressive visits}

\begin{figure}[t]
\centering
\includegraphics[width=0.6\textwidth]{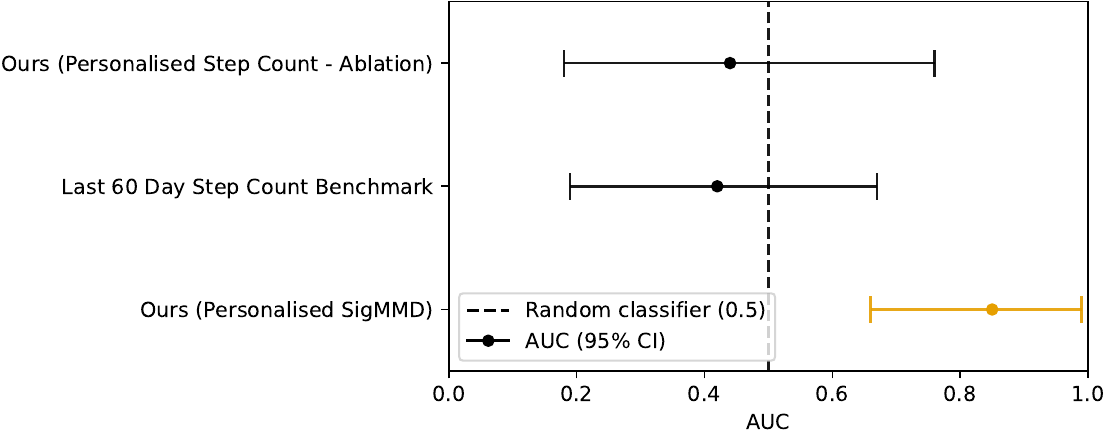}
\caption{Comparison of AUC (with 95\% confidence intervals) for the personalised SigMMD model against the step count benchmark and ablation model. Both step count approaches perform below a random classifier (AUC = 0.5, dashed line), with confidence intervals that cross it, whereas the personalised SigMMD model achieves a substantially higher AUC.}
\label{fig:auc_ci_comparison}
\end{figure}

As an ablation of the personalised SigMMD model, we implemented an analogous step count model that replaces the SigMMD feature with absolute deviation between daily steps in the comparison window and the mean daily steps in the baseline set. This isolates whether detection of progression stems from the shape of the time series captured by SigMMD, or merely from changes in overall daily activity level. The step-count ablation achieved a lower performance than the personalised SigMMD model. Under the same normalisation and the same permutation test, the step-count ratios were not higher at progressive visits (p-value = 0.61 over 4,320 permutations). To confirm that this result was not an artefact of our particular ablation design, we further benchmarked against a more conventional step-count classification model using daily walking step counts extracted from the accelerometer data over the 60 days preceding each clinic visit. We report classification metrics with 95\% confidence intervals for both step-count models and the personalised SigMMD model in \Cref{tab:classification_metrics}, with AUC confidence intervals visualised in \Cref{fig:auc_ci_comparison}. In contrast to personalised SigMMD, neither step-count model showed clear evidence of discriminative ability beyond chance. This suggests that mean daily step count is an insufficient summary measure for detecting disease progression from accelerometry data, warranting the additional complexity introduced by SigMMD.

\section{Conclusion and Limitations}
\label{sec:conclusion}
In this work, we introduced a novel application of rough path theory to track disease progression in patients with HGG using wearable accelerometry. We demonstrated that personalised SigMMD, a principled, training-free patient-specific metric, effectively captures the divergence in walking path distributions associated with tumour progression. Evaluated on the BrainWear dataset, this single-feature signature-based classifier achieved an AUC of $0.8545$ using the ENMO representation, outperforming an interpretable clinical benchmark model trained using raw step counts. Our results highlight the advantages of rough path signatures for noisy, label-scarce time series data. 

While our findings establish a proof-of-concept, we note some limitations to contextualise these exploratory results. Our evaluation is constrained by data scarcity. The personalised normalisation requires each patient to have at least one stable clinic window, limiting our evaluation cohort to 26 clinic visits across 11 patients. The normalisation also means that the personalised ratio cannot yet be computed prospectively. A patient's stable reference uses that patient's visits labelled stable, which can include visits after a progression. Our analysis therefore shows that a patient's walking differs between stable and progressive periods, but leaves open whether progression can be detected before it is diagnosed. With a larger longitudinal cohort, the reference should instead be computed only from stable visits that precede the visit being scored.

The development of mathematically principled digital biomarkers can have profound implications for neuro-oncology. By providing a scalable, non-invasive mechanism to quantify functional status in real-world environments, this framework could complement intermittent clinical assessments and support more timely, data-driven intervention. To advance this proof-of-concept, an important next step is to move from clinic-level aggregation towards continuous modelling of daily SigMMD trajectories. This would enable investigation of real-time shifts in walking dynamics and, ultimately, whether such changes can identify the onset of functional decline.

\newpage
\section*{Acknowledgements}
Barbora Barancikova is supported by UK Research and Innovation [UKRI Centre for Doctoral Training in AI for Healthcare grant number EP/S023283/1]. 

\bibliography{bibliography}
\bibliographystyle{unsrtnat}

\newpage

\appendix

\section{Additional Results}
\subsection{Triaxial Representation}\label{sec:triaxial}

\begin{figure}[ht]
\centering
\begin{subfigure}{0.4\textwidth}
\centering
\includegraphics[width=0.9\linewidth]{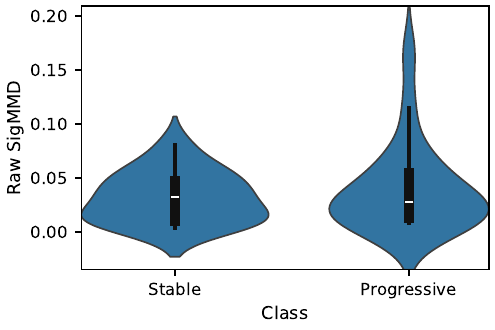}
\caption{Raw mean SigMMD.}
\end{subfigure}
\begin{subfigure}{0.4\textwidth}
\centering
\includegraphics[width=0.9\linewidth]{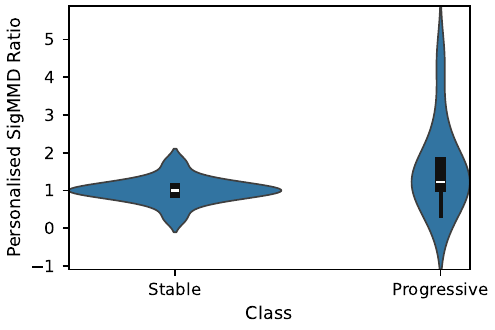}
\caption{Personalised SigMMD ratio.}
\end{subfigure}
\caption{Distributions of raw mean SigMMD values and personalised SigMMD ratios for stable and progressive-disease visits, for the Triaxial representation.}
\label{fig:violin_plots_tri}
\end{figure}

We also repeated the experiment in \Cref{subsec:persResults} for the raw triaxial accelerometer data representation instead of ENMO. The experimental setting remained the same, except each path in the baseline and comparison set had 4 channels. Similar to the ENMO representation, raw SigMMDs show significant overlap unlike the personalised SigMMDs as shown in \Cref{fig:violin_plots_tri}. Personalised ratios were again higher at progressive than at stable visits (within-patient permutation test, p-value = 0.037 over 4,320 permutations).

\begin{figure}[ht]
\centering
\begin{subfigure}{0.48\textwidth}
    \centering
    \includegraphics[width=0.7\linewidth]{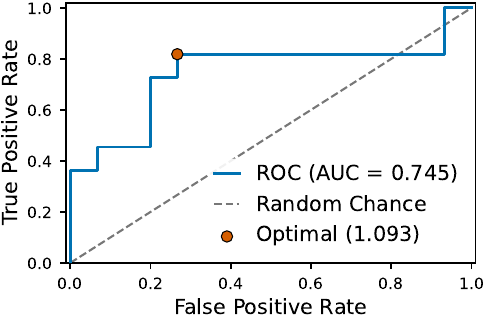}
    \caption{ROC curve generated using personalised SigMMDs from the triaxial representation.}
\end{subfigure}
\hfill
\begin{subfigure}{0.48\textwidth}
    \centering
    \includegraphics[width=0.7\linewidth]{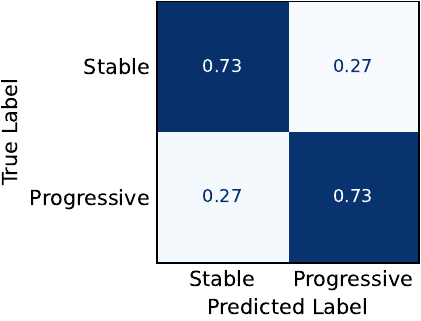}
    \caption{Confusion matrix generated by Leave One Patient Out Classification from triaxial representation.}
\end{subfigure}
\caption{Performance evaluation of the triaxial representation using an ROC curve and confusion matrix.}
\label{fig:conf_mat_tri}
\end{figure}

An ROC curve generated on personalised SigMMD ratio thresholds showed an AUC of $0.745$ (\Cref{fig:conf_mat_tri}). A confusion matrix was computed from results generated by LOPO classification (\Cref{fig:conf_mat_tri}). Full performance metrics are reported in \Cref{tab:classification_metrics_tri}.

\begin{table}[h!]
\caption{Classification Performance generated by Leave One Patient Out Classification for Triaxial representation.}
\centering
\resizebox{\textwidth}{!}{
\begin{tabular}{lccccc}
\toprule
\textbf{Model} & \textbf{AUC} & \textbf{Sensitivity} & \textbf{Specificity} & \textbf{Precision} \\
\midrule
Personalised SigMMD & 0.7455 (0.5125-0.9546) & 0.7272 (0.400-1.000)  & 0.7333 (0.6111-0.9167) & 0.6667 (0.4000-0.8750)\\
\bottomrule
\end{tabular}
}
\label{tab:classification_metrics_tri}
\end{table}

\subsection{Kolmogorov-Smirnov Test} \label{subsec:ksResults}

Here, we report the full results for the experiment run in \Cref{subsec:activitydiff}. We use signature coefficients up to truncation level 2.  \Cref{tab:ks_enmo} and \Cref{tab:ks_tri} show the results of the KS test between \textit{walking} and other label distributions for all non-trivial signature coefficients, for the ENMO and triaxial representations respectively. The triaxial representation also showed all non-trivial signature coefficients were statistically significant apart from some in \textit{walking vs mixed} label comparison. With time augmentation, the ENMO representation has 2 dimensions leading to 6 signature coefficients while the Triaxial representation has 4 dimensions leading to 20 signature coefficients.

\begin{table}[h!]
\centering
\caption{KS Test Results for ENMO representation}
\begin{tabular}{lcrrc}
\toprule
\textbf{Comparison} & \textbf{Signature Coefficient} & \textbf{ks stat} & \textbf{p-value} & \textbf{Significant} \\
\midrule
\multirow{4}{*}{walking vs mixed}
    & $S_1(x)^2$    & 0.064 & 0.033253 & True \\
    & $S_2(x)^{1,2}$  & 0.117 & 2.21E-06   & True \\
    & $S_2(x)^{2,1}$  & 0.106 & 2.60E-05   & True \\
    & $S_2(x)^{2,2}$  & 0.108 & 1.69E-05   & True \\
\midrule
\multirow{4}{*}{walking vs sleep}

   & $S_1(x)^2$    & 0.223 & 3.41E-22   & True \\
   & $S_2(x)^{1,2}$  & 0.148 & 5.73E-10   & True \\
   & $S_2(x)^{2,1}$  & 0.218 & 3.20E-21   & True \\
   & $S_2(x)^{2,2}$  & 0.253 & 1.63E-28   & True \\
\midrule
\multirow{4}{*}{walking vs vehicle}

 & $S_1(x)^2$    & 0.218 & 3.20E-21   & True \\
 & $S_2(x)^{1,2}$  & 0.184 & 3.32E-15   & True \\
 & $S_2(x)^{2,1}$  & 0.137 & 1.34E-08   & True \\
 & $S_2(x)^{2,2}$  & 0.331 & 6.84E-49   & True \\
\midrule
\multirow{4}{*}{walking vs sit-stand}

 & $S_1(x)^2$   & 0.368 & 1.30E-60   & True \\
 & $S_2(x)^{1,2}$ & 0.298 & 1.45E-39   & True \\
 & $S_2(x)^{2,1}$ & 0.317 & 8.32E-45   & True \\
 & $S_2(x)^{2,2}$ & 0.555 & 5.25E-142  & True \\
\bottomrule
\end{tabular}
\label{tab:ks_enmo}
\end{table}
\begin{longtable}{lcrrc}
\caption{KS Test Results for Triaxial representation}\\
\toprule
\textbf{Comparison} & \textbf{Signature Coefficient} & \textbf{ks stat} & \textbf{p-value} & \textbf{Significant} \\
\midrule
\endfirsthead

\caption*{KS Test Results for Triaxial representation (continued)}\\
\toprule
\textbf{Comparison} & \textbf{Signature Coefficient} & \textbf{ks stat} & \textbf{p-value} & \textbf{Significant} \\
\midrule
\endhead

\bottomrule
\endfoot

\multirow{18}{*}{walking vs mixed}
 & $S_1(x)^2$   & 0.06  & 0.054627   & False \\
 & $S_1(x)^3$   & 0.055 & 0.097103   & False \\
 & $S_1(x)^4$   & 0.031 & 0.722825   & False \\
 & $S_2(x)^{1,2}$ & 0.086 & 0.001221   & True  \\
 & $S_2(x)^{1,3}$ & 0.091 & 0.000503   & True  \\
 & $S_2(x)^{1,4}$ & 0.084 & 0.001716   & True  \\
 & $S_2(x)^{2,1}$ & 0.089 & 0.000721   & True  \\
 & $S_2(x)^{2,2}$ & 0.057 & 0.077611   & False \\
 & $S_2(x)^{2,3}$ & 0.202 & 2.93E-18   & True  \\
 & $S_2(x)^{2,4}$ & 0.208 & 2.42E-19   & True  \\
 & $S_2(x)^{3,1}$ & 0.096 & 0.000197   & True  \\
 & $S_2(x)^{3,2}$ & 0.201 & 4.41E-18   & True  \\
 & $S_2(x)^{3,3}$ & 0.094 & 0.000288   & True  \\
 & $S_2(x)^{3,4}$ & 0.152 & 1.71E-10   & True  \\
 & $S_2(x)^{4,1}$ & 0.081 & 0.002818   & True  \\
 & $S_2(x)^{4,2}$ & 0.221 & 8.40E-22   & True  \\
 & $S_2(x)^{4,3}$ & 0.156 & 4.94E-11   & True  \\
 & $S_2(x)^{4,4}$ & 0.055 & 0.097103   & False \\
\midrule

\pagebreak
\multirow{18}{*}{walking vs sleep}
 & $S_1(x)^2$   & 0.135 & 2.32E-08   & True  \\
 & $S_1(x)^3$   & 0.156 & 4.94E-11   & True  \\
 & $S_1(x)^4$   & 0.108 & 1.69E-05   & True  \\
 & $S_2(x)^{1,2}$ & 0.154 & 9.23E-11   & True  \\
 & $S_2(x)^{1,3}$ & 0.19  & 3.43E-16   & True  \\
 & $S_2(x)^{1,4}$ & 0.146 & 1.04E-09   & True  \\
 & $S_2(x)^{2,1}$ & 0.148 & 5.73E-10   & True  \\
 & $S_2(x)^{2,2}$ & 0.224 & 2.16E-22   & True  \\
 & $S_2(x)^{2,3}$ & 0.368 & 1.30E-60   & True  \\
 & $S_2(x)^{2,4}$ & 0.367 & 2.81E-60   & True  \\
 & $S_2(x)^{3,1}$ & 0.142 & 3.30E-09   & True  \\
 & $S_2(x)^{3,2}$ & 0.359 & 1.23E-57   & True  \\
 & $S_2(x)^{3,3}$ & 0.281 & 3.63E-35   & True  \\
 & $S_2(x)^{3,4}$ & 0.291 & 1.01E-37   & True  \\
 & $S_2(x)^{4,1}$ & 0.097 & 0.000162   & True  \\
 & $S_2(x)^{4,2}$ & 0.377 & 1.14E-63   & True  \\
 & $S_2(x)^{4,3}$ & 0.297 & 2.68E-39   & True  \\
 & $S_2(x)^{4,4}$ & 0.184 & 3.32E-15   & True  \\
\midrule

\multirow{18}{*}{walking vs vehicle}
 & $S_1(x)^2$   & 0.116 & 2.80E-06   & True  \\
 & $S_1(x)^3$   & 0.148 & 5.73E-10   & True  \\
 & $S_1(x)^4$   & 0.098 & 0.000133   & True  \\
 & $S_2(x)^{1,2}$ & 0.173 & 1.75E-13   & True  \\
 & $S_2(x)^{1,3}$ & 0.2   & 6.61E-18   & True  \\
 & $S_2(x)^{1,4}$ & 0.173 & 1.75E-13   & True  \\
 & $S_2(x)^{2,1}$ & 0.157 & 3.61E-11   & True  \\
 & $S_2(x)^{2,2}$ & 0.222 & 5.36E-22   & True  \\
 & $S_2(x)^{2,3}$ & 0.256 & 3.42E-29   & True  \\
 & $S_2(x)^{2,4}$ & 0.228 & 3.44E-23   & True  \\
 & $S_2(x)^{3,1}$ & 0.136 & 1.77E-08   & True  \\
 & $S_2(x)^{3,2}$ & 0.243 & 2.58E-26   & True  \\
 & $S_2(x)^{3,3}$ & 0.214 & 1.85E-20   & True  \\
 & $S_2(x)^{3,4}$ & 0.159 & 1.91E-11   & True  \\
 & $S_2(x)^{4,1}$ & 0.146 & 1.04E-09   & True  \\
 & $S_2(x)^{4,2}$ & 0.244 & 1.57E-26   & True  \\
 & $S_2(x)^{4,3}$ & 0.168 & 9.80E-13   & True  \\
 & $S_2(x)^{4,4}$ & 0.182 & 6.95E-15   & True  \\
\midrule

\multirow{18}{*}{walking vs sit-stand}
 & $S_1(x)^2$   & 0.241 & 6.91E-26   & True  \\
 & $S_1(x)^3$   & 0.204 & 1.29E-18   & True  \\
 & $S_1(x)^4$   & 0.196 & 3.29E-17   & True  \\
 & $S_2(x)^{1,2}$ & 0.275 & 1.12E-33   & True  \\
 & $S_2(x)^{1,3}$ & 0.239 & 1.84E-25   & True  \\
 & $S_2(x)^{1,4}$ & 0.254 & 9.71E-29   & True  \\
 & $S_2(x)^{2,1}$ & 0.211 & 6.75E-20   & True  \\
 & $S_2(x)^{2,2}$ & 0.451 & 5.57E-92   & True  \\
 & $S_2(x)^{2,3}$ & 0.413 & 1.00E-76   & True  \\
 & $S_2(x)^{2,4}$ & 0.427 & 3.62E-82   & True  \\
 & $S_2(x)^{3,1}$ & 0.188 & 7.37E-16   & True  \\
 & $S_2(x)^{3,2}$ & 0.41  & 1.38E-75   & True  \\
 & $S_2(x)^{3,3}$ & 0.391 & 1.37E-68   & True  \\
 & $S_2(x)^{3,4}$ & 0.336 & 2.14E-50   & True  \\
 & $S_2(x)^{4,1}$ & 0.196 & 3.29E-17   & True  \\
 & $S_2(x)^{4,2}$ & 0.435 & 2.27E-85   & True  \\
 & $S_2(x)^{4,3}$ & 0.342 & 3.10E-52   & True  \\
 & $S_2(x)^{4,4}$ & 0.372 & 5.82E-62   & True  \\
\label{tab:ks_tri}
\end{longtable}

\section{Compute Resources}
\label{sec:compute}

All experiments were run on AMD EPYC 7742 nodes with 64 GB RAM. Accelerometry preprocessing and SigMMD computation, for which we used the Python pysiglib~\cite{Shmelev_pySigLib--Fast_Signature-Based_Computations_2025} toolkit, accounted for nearly all compute, requiring approximately 6 hours per patient in total; model fitting and evaluation were negligible by comparison. The experiments are reproducible on any standard modern CPU, although GPU implementations of SigMMD and preprocessing would likely substantially reduce runtime.

\end{document}